\documentclass[conference,9pt]{IEEEtran}
\IEEEoverridecommandlockouts
\usepackage{colortbl}
\usepackage{hyperref}
\hypersetup{hypertex=true,
            colorlinks=true,
            linkcolor=red,
            anchorcolor=blue,
            citecolor=green}
\usepackage{cite}
\usepackage{amsmath,amssymb,amsfonts}
\usepackage{algorithm}
\usepackage{algorithmic}
\usepackage{graphicx}
\usepackage{textcomp}
\usepackage{xcolor}
\usepackage{setspace}
\usepackage{bbding}
\usepackage{arydshln}
\usepackage{multirow}
\usepackage{makecell}
\usepackage{booktabs}
\usepackage{nicematrix}

\newcommand{\firstcolor}[1]{\colorbox{red!15}{#1}}
\newcommand{\secondcolor}[1]{\colorbox{yellow!40}{#1}}
\newcommand{\thirdcolor}[1]{\colorbox{blue!15}{#1}}

\definecolor{lightred}{RGB}{255, 102, 102}
\newcommand{\mycolor}[2][1]{
    \textcolor[rgb]{0,0,#1}{#2}
}

\def\BibTeX{{\rm B\kern-.05em{\sc i\kern-.025em b}\kern-.08em
    T\kern-.1667em\lower.7ex\hbox{E}\kern-.125emX}}
\begin{document}

\title{CAE-DFKD: Bridging the Transferability Gap in Data-Free Knowledge Distillation
\vspace{-0.5cm}
}

\DeclareRobustCommand*{\IEEEauthorrefmark}[1]{%
  \raisebox{0pt}[0pt][0pt]{\textsuperscript{\footnotesize #1}}%
}

\author{
\thanks{Zherui Zhang: zzr787906410@bupt.edu.cn}
    \IEEEauthorblockN{
        Zherui Zhang\IEEEauthorrefmark{1}$^{*}$, 
        Changwei Wang\IEEEauthorrefmark{3,4}$^{*}$, 
        Rongtao Xu\IEEEauthorrefmark{2}, 
        Wenhao Xu\IEEEauthorrefmark{1},
        Shibiao Xu\IEEEauthorrefmark{1}$^{\dagger}$\thanks{$^{\dagger}$ Shibiao Xu is the corresponding author: shibiaoxu@bupt.edu.cn},
        Yu Zhang\IEEEauthorrefmark{5},
        Jie Zhou\IEEEauthorrefmark{1} and 
        Li Guo\IEEEauthorrefmark{1}
        \thanks{$^{*}$ Zherui Zhang and Changwei Wang contributed equally.}
    }
    \IEEEauthorblockA{
        \IEEEauthorrefmark{1} School of Artificial Intelligence, Beijing University of Posts and Telecommunications, Beijing, China \\
        \IEEEauthorrefmark{2} The State Key Laboratory of Multimodal Artificial Intelligence Systems, Institute of Automation, \\ Chinese Academy of Sciences, Beijing, China \\
        \IEEEauthorrefmark{3} The Key Laboratory of Computing Power Network and Information Security, Ministry of Education, \\ Shandong Computer Science Center, Qilu University of Technology, Jinan, China. \\
        \IEEEauthorrefmark{4} Shandong Provincial Key Laboratory of Computing Power Internet and Service Computing, \\ Shandong Fundamental Research Center for Computer Science, Jinan, China \\
       \IEEEauthorrefmark{5} Tongji University, Shanghai, China \\
    }
    
    \vspace{-1cm}
}

\maketitle

\begin{abstract}
Data-Free Knowledge Distillation (DFKD) enables the knowledge transfer from the given pre-trained teacher network to the target student model without access to the real training data.
Existing DFKD methods focus primarily on improving image recognition performance on associated datasets, often neglecting the crucial aspect of the transferability of learned representations. 
In this paper, we propose Category-Aware Embedding Data-Free Knowledge Distillation (CAE-DFKD), which addresses at the embedding level the limitations of previous rely on image-level methods to improve model generalization but fail when directly applied to DFKD. The superiority and flexibility of CAE-DFKD are extensively evaluated, including:
\textit{\textbf{i.)}} Significant efficiency advantages resulting from altering the generator training paradigm;
\textit{\textbf{ii.)}} Competitive performance with existing DFKD state-of-the-art methods on image recognition tasks;
\textit{\textbf{iii.)}} Remarkable transferability of data-free learned representations demonstrated in downstream tasks.
\end{abstract}

\begin{IEEEkeywords}
Knowledge Distillation, Representation Learning, Transfer Learning, Contrastive Learning.
\end{IEEEkeywords}

\section{Introduction}
\label{Introduction}
\begin{figure}[!t]
\setlength{\abovecaptionskip}{-0.2cm}
\begin{center}
  \includegraphics[width=0.8\linewidth]{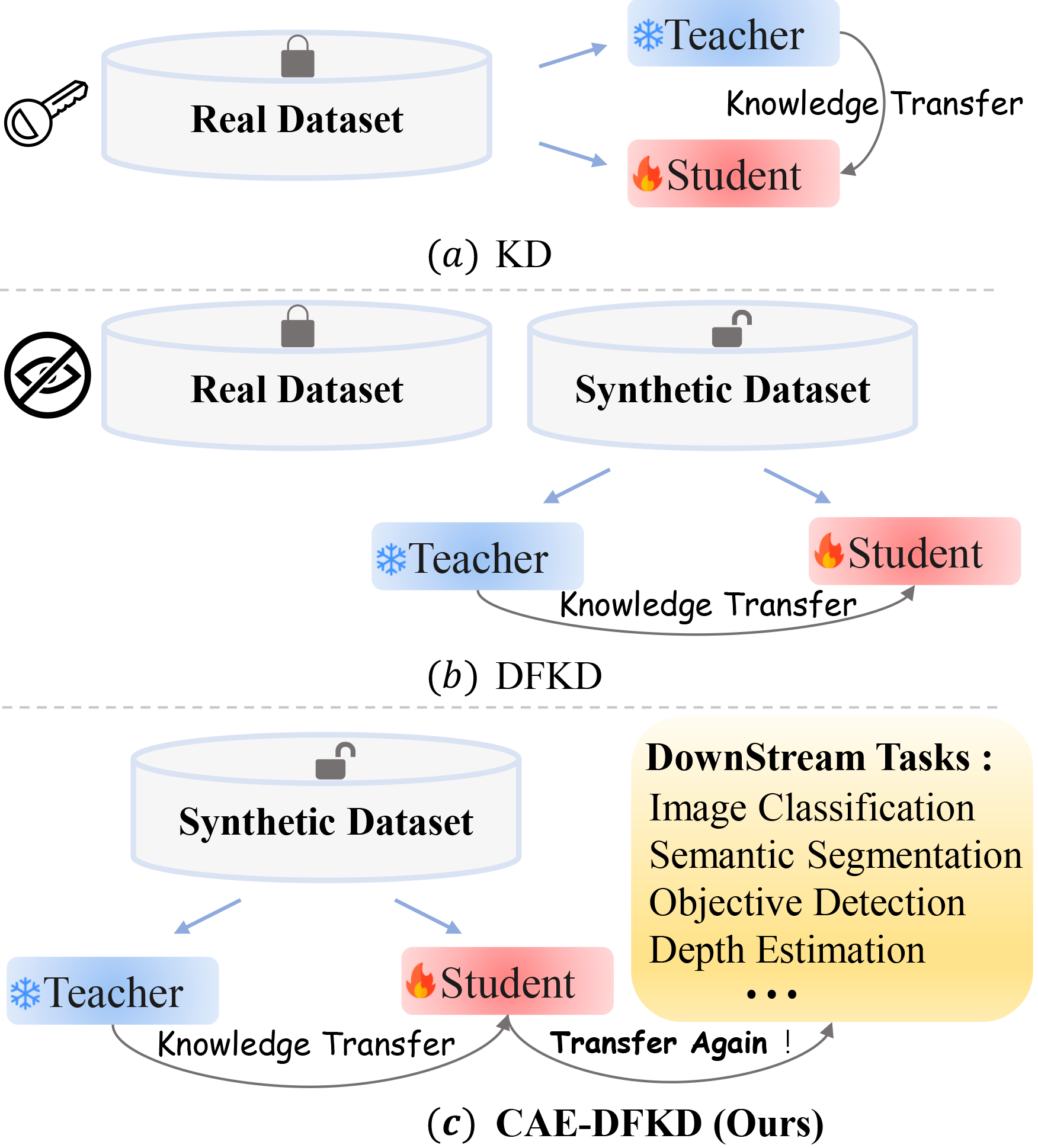}
\end{center}
   \caption{
   \textbf{From KD to CAE-DFKD.}
$(a)$ The traditional paradigm of knowledge distillation~(KD) has access to the real dataset.
$(b)$ Data-Free Knowledge Distillation (DFKD) is unaware of the read dataset and employs synthetic datasets for knowledge transfer.
$(c)$ The goal of CAE-DFKD is to continually transfer the knowledge acquired under data-free setting to downstream tasks.
}
\label{fig:kdtoCAE-DFKD}
\vspace{-0.5cm}
\end{figure}

\IEEEPARstart{K}{nowledge} distillation (KD) transfers knowledge from high-capacity teacher networks to compact student networks, enabling comparable performance with reduced computational costs.  
Although effective for various computer vision tasks (e.g., image classification~\cite{guo2023class}, object detection~\cite{wang2024crosskd,bang2024radardistill}, and semantic segmentation~\cite{yuan2024fakd,li2024knowledge}), traditional KD requires access to real training dataset (Figure \ref{fig:kdtoCAE-DFKD}~\textbf{(a)}), presenting privacy concerns. 
Data-free knowledge distillation (DFKD)~\cite{lopes2017data} addresses this by reconstructing training data from the given pre-trained teacher network (Figure \ref{fig:kdtoCAE-DFKD}~\textbf{(b)}), are broadly categorized into optimization-based~\cite{jiang2023delving,cazenavette2022dataset} and generator-based views.  
This paper focuses on generator-based DFKD~\cite{wang2023sampling,patel2023learning,frank2023data,hao2022cdfkd}, where a trainable generator network synthesizes proxy dataset guided by the pre-trained teacher that as the non-updatable discriminator.

\begin{figure*}[tb]
    \centering
    \includegraphics[width=0.9\textwidth]{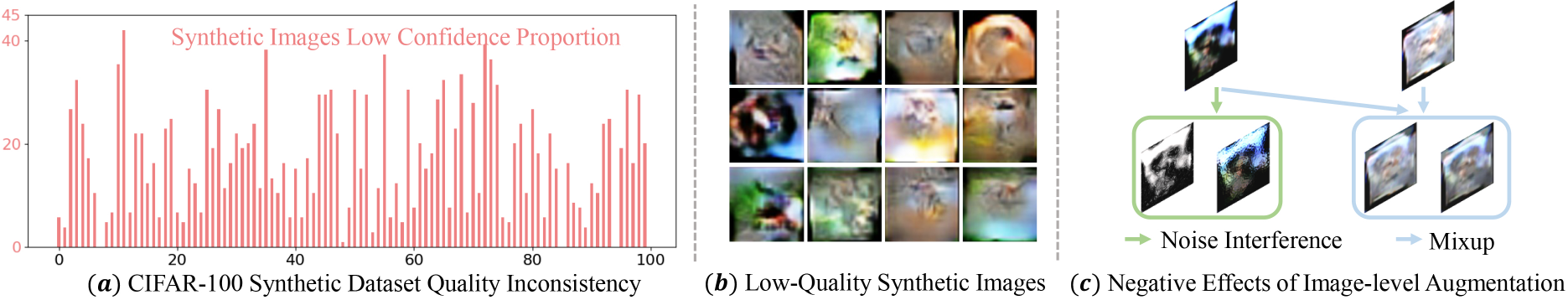}
    \caption{
    \textbf{Quality Difference in Synthetic Images.}
$(a)$ The proportion of low-confidence ($\leq 0.1$ highest probability) synthetic images within the corresponding categories. $(b)$ The generator inevitably produces semantically ambiguous, low-quality synthetic  images. $(c)$ Applying image-level noise interference or high-intensity data augmentation to uncertain images renders their semantics more abstract and difficult to exploit.
    }
    \label{fig:weak}
    \vspace{-0.5cm}
\end{figure*}

Despite advances in generator-based DFKD for image recognition~\cite{nayer,yu2023data}, a key research gap remains to understand and improve the transferability of learned representations to new downstream tasks and datasets.
Furthermore, existing DFKD explorations also suffer from two key limitations: \textit{\textbf{i)}} the teacher category prediction preferences introduce distributional shifts in the synthetic dataset, leading to inconsistent synthetic image quality across categories~\cite{wang2024confounded,wang2024sampling,li2024towards}; \textit{\textbf{ii)}} the generator network training process can suffer from pattern collapse, leading to numerous noisy or worthless synthetic images. 
As shown in Figure \ref{fig:weak}~\textbf{(a)}, the synthetic images on CIFAR-100 exhibits significant  difference across categories in low-confidence proportion from the teacher view~($\leq 0.1$ highest probability), indicating inconsistent synthetic image quality and potential category imbalance.  
Figure~\ref{fig:weak}~\textbf{(b)} visualizes the lack of clear semantic content in low-confidence synthetic images. 
Consequently, existing methods for improving model generalization or task adaptability, such as high-intensity data augmentation or auxiliary supervised objectives, directly applying methods such as Mixup or contrastive learning to synthetic images, can negatively impact student network performance (Table~\ref{tab:ev1}), as transformations applied to semantically ambiguous synthetic images further degrade their usefulness (Figure \ref{fig:weak}~\textbf{(c)}).

\begin{table}[!t]
    \centering
    \caption{
    Directly applying contrastive learning or Mixup, which are designed for image-level, to DFKD synthetic images yields adverse effects. 
    Setting: CIFAR-100, ResNet-34 $\rightarrow$ ResNet-18.
    }
    \setlength{\tabcolsep}{4mm}{
    \resizebox{0.7\linewidth}{!}{
        \begin{tabular}{ll}
            \toprule[1.5px] 
            \textbf{Method} &  \textbf{Top-1 Acc.(\%)} \\
            \hline
            Vanilla & 77.54 \\
            \hline
            \color{red}{+ }\color{black} Mixup & 77.13\color{red}$\downarrow$\color{black} \\
            \color{red}{+ }\color{black} Contrastive Learning & 76.38\color{red}$\downarrow$\color{black} \\
            \bottomrule[1.5px] 
        \end{tabular}
    }
    }
    \label{tab:ev1}
    \vspace{-0.4cm}
\end{table}

This paper addresses these limitations by shifting from image-level operation to embedding-level operation for DFKD, promoting more robust and transferable feature learning. By operating in the embedding space, our \textbf{{{\large {C}}}ategory-{{\large {A}}}ware {{\large {E}}}mbedding Data-Free Knowledge Distillation (CAE-DFKD)} method mitigates the adverse effects of quality inconsistencies in synthetic images, and contributions are as follows:
\begin{itemize}
    \item \textbf{Category Embedding Noise Diffusion (CEND) Layer}: Unlike native DFKD setting that sample generator embedding inputs from unstructured Gaussian noise, we introduce a structured embedding space with category priors using pre-trained language models. The proposed CEND layer applies multi-source noise to dynamically diffuse category embeddings, enriching the structured embedding space, and contributes to the fast convergence of the generator.
    \item \textbf{Category Noise Contrastive Learning (CNCL)}: CNCL is an embedding-level contrastive learning method designed for DFKD that exploits the generator network to \emph{indirectly} produce positive-negative contrastive pairs, encouraging the student network to focus on domain-invariant category features with stronger generalization.
    \item \textbf{Effective Data-Free Transferability}: CAE-DFKD integrates the advantages of the CEND and CNCL components, enabling effective data-free knowledge transfer to various downstream tasks~(Figure \ref{fig:kdtoCAE-DFKD}~(c)), surpassing the adaptability and practicality of existing DFKD methods.
\end{itemize}

\section{Related Work}

\subsubsection{\texttt{\textbf{Data-free KD~(DFKD)}}}
DFKD transfers knowledge from the given pre-trained teacher to the target student network using synthetic images, avoiding access to the real training dataset.  
Generator-based DFKD methods~\cite{fang2022up,nayer,yu2023data,wang2024confounded}, the focus of this paper, adversarially train a generator network and a student network under the guidance of the given teacher network that as the un-trainable discriminator.
The generator synthesizes images to maximize the divergence between student and teacher predictions, while the student learns to align its predictions with the teacher in these synthetic images~\cite{fang2022up, wang2023sampling, frank2023data}.

However, generator-based DFKD faces a key challenge: synthetic dataset quality. 
These generated images often do not fully capture the diversity and complexity of real data, and their quality is heavily influenced by the teacher network prediction preferences~\cite{wang2024confounded, wang2024sampling}. 
Existing DFKD methods attempt to address this limitation through various strategies.  Some optimize generator features to promote diversity (e.g. SpaceShipNet~\cite{yu2023data}), while others incorporate causal inference (KDCI~\cite{wang2024confounded}), leverage open-world data (ODSD~\cite{wang2024sampling}), implement curriculum learning (AdaDFKD~\cite{li2024adadfkd}), or employ diffusion-based augmentation (DDA~\cite{li2024towards}). 
NAYER~\cite{nayer} periodically re-initializes the generator for improved diversity.

Although these methods improve image recognition performance on the corresponding real dataset by improving synthetic image quality, their effectiveness on broader downstream tasks remains largely unexplored.  
In other words, it remains unknown whether the representations learned following the data-free setting possess transferability.
This paper fills this gap by introducing an embedding-level paradigm to DFKD, enabling data-free and privacy-preserving knowledge transfer for a wider range of real-world applications.

\subsubsection{\texttt{\textbf{Contrastive Learning}}}
\label{Self-supervised Learning}
Self-supervised learning exploits pretext tasks on unlabeled data to obtain robust and discriminative representations~\cite{li2022fakeclr, ci2022fast, yang2022concl, yuan2022pointclm}.
Among these pretext tasks, contrastive learning~\cite{liu2023hard, denize2023similarity,tu2023hierarchically, zhu2023hnssl} outperforms many downstream tasks by bringing semantically similar data closer to the latent space while pushing apart differing data. However, the effectiveness of contrastive learning is highly dependent on how positive and negative pairs are defined\cite{liu2023hard}.
Conventional methods generally produce these contrastive pairs at the image level \cite{yang2022concl,li2022fakeclr}, but under the DFKD setting, this strategy risks introducing greater noise or more marked semantic ambiguity, thus increasing uncertainty in the optimization objective of the student network, as discussed in Section~\ref{Introduction}.

In this paper, we exploit the synthesizing ability of the generator-based DFKD to address this issue. 
Instead of constructing positive-negative contrastive pairs at the image level, we introduce embedding-level design that avoids the potential degradation resulting from low-quality synthetic images, allowing the student model to focus on more generalizable domain-invariant features.

\section{Method}

\begin{figure*}[t]
\centering
\setlength{\abovecaptionskip}{-0.2cm}
\includegraphics[width=0.9\linewidth]{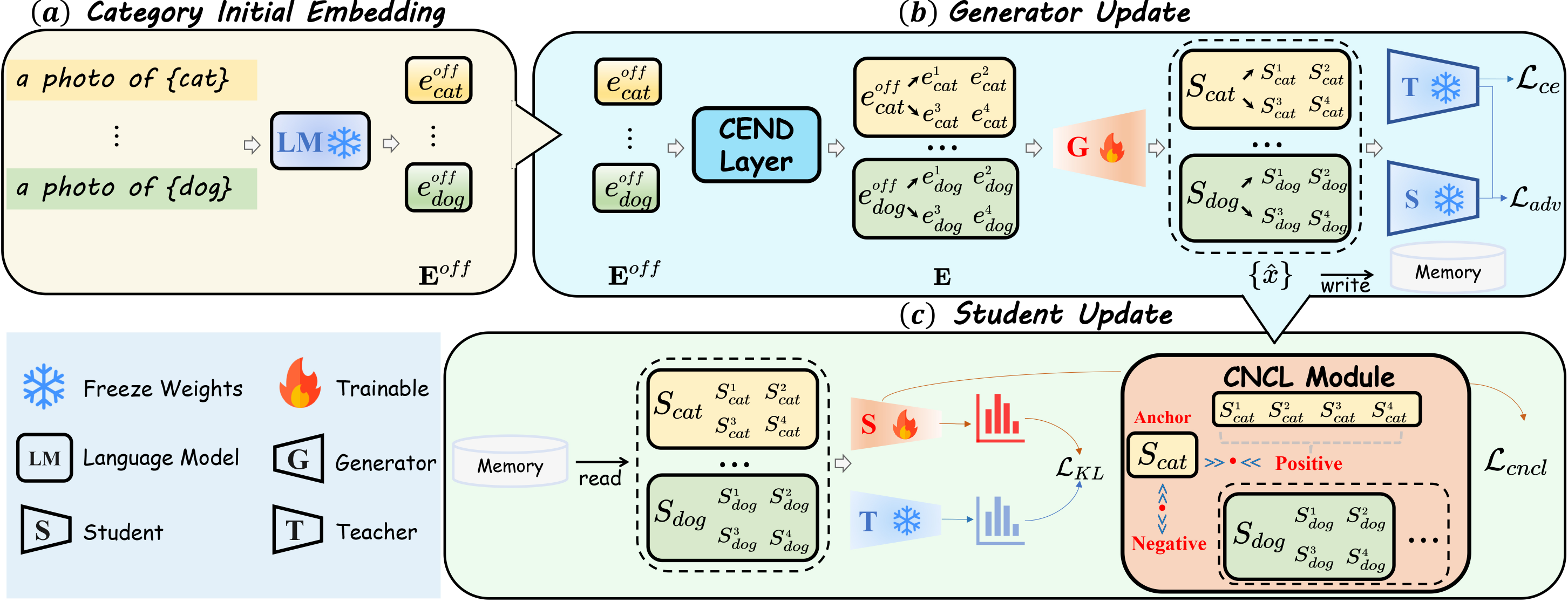}
\caption{\textbf{CAE-DFKD Framework}.
(a)  A pre-trained language model (LM, default CLIP) provides a category-structured initial embedding space, $\mathbf{E}^{\text{off}}$, contrasting with the un-structured Gaussian noise used in native DFKD, this process is performed offline and does not cause any training burden;
(b) During generator updates, the Category Embedding Noise Diffusion (CEND) layer addresses $\mathbf{E}^{\text{off}}$ sparsity and lack of diversity. CEND samples from noise sources with distinct pre-set distributions, dynamically diffusing category embeddings to induce a diverse embedding space, $\mathbf{E}$. synthetic images, written to memory, are fed to teacher and student networks to compute cross-entropy ($\mathcal{L}_{ce}$), adversarial ($\mathcal{L}_{adv}$), and batch normalization ($\mathcal{L}_{bn}$) losses for generator updates;
(c) During student updates, classic logit knowledge distillation ($\mathcal{L}_{kl}$) is performed on synthetic images read from memory. Further, Category Noise Contrastive Learning (CNCL) constructs embedding-level positive-negative pairs, optimizing towards $\mathcal{L}_{cncl}$.
}

\label{fig:f3}
\vspace{-0.4cm}
\end{figure*}

\subsection{Problem Formulation and Overview}

\subsubsection{\texttt{\textbf{KD $\rightarrow$ DFKD}}}
Given a training dataset $\mathcal{D} = \{(x_i, y_i)\}_{i=1}^{\mathcal{I}}$, where $\mathcal{I}$ represents the length of $\mathcal{D}$, and each image $x_i \in \mathbb{R}^{3 \times h \times w}$, where $h \times w$ are the input resolution, $y_i \in \{1, 2, \cdots, K\}$ denotes its corresponding hard label. 
Knowledge distillation (KD) aims to transfer knowledge from the given pre-trained teacher network $\mathcal{T}(x; \theta_T)$ to the smaller target student network $\mathcal{S}(x; \theta_S)$. 
The objective is for the student to approximate the teacher network output logits on $\mathcal{D}$ through the following optimization:
\begin{equation}
    \begin{aligned}
    \min_{{\theta}_{s}} \mathbb{E}_{(x_i,y_i) \in \mathcal{D}} \Big[ \mathcal{L}_{\text{KL}} 
    &\left( \mathcal{S}(x_i;\theta_{S}), \mathcal{T}(x_i;\theta_{T}) \right) \\
    &+ \lambda \mathcal{L}_{\text{CE}} \left( \mathcal{S}(x_i;\theta_{S}), y_i \right) \Big],
    \end{aligned}
\end{equation}
where $\lambda$ is used as a hyperparameter to balance the Kullback-Leibler divergence~($\mathcal{L}_{KL}$) and cross-entropy loss~($\mathcal{L}_{CE}$).

By comparison, data-free knowledge distillation (DFKD) seeks to train the student network without access to the real training dataset $\mathcal{D}$. 
Generator-based DFKD achieves this by employing adversarial training, in which a generator $G$ produces synthetic images used to train the student:
\begin{equation}
\begin{aligned}
  \min_{\theta_S} \max_{\theta_G} \mathbb{E}_{z \sim \mathcal{N}(0, 1)} \left[ \mathcal{L}_{KL} \left( \mathcal{S}(\hat{x}; \theta_S), \mathcal{T}(\hat{x}; \theta_T) \right) \right],
\end{aligned}
\label{eq:adv}
\end{equation}
where $\hat{x} = G(z, y; \theta_G)$ is a synthetic image produced by the generator $G$, and $z$ is embedding vector sampled from the Gaussian distribution.

\subsubsection{\texttt{\textbf{Motivation}}}
The goal of existing DFKD methods is to improve image recognition performance on the corresponding real dataset, yet there is insufficient exploration on whether the performance gains obtained through data-free paradigm can benefit other downstream tasks.

\subsubsection{\texttt{\textbf{CAE-DFKD Framework}}}
Figure~\ref{fig:f3} demonstrates how our proposed CAE-DFKD flexibly incorporates into the existing DFKD framework. 
During the generator network update, we replace the native unstructured Gaussian noise sampling with the initial category embedding $\mathbf{E}^{\text{off}}$, and employ the Category Embedding Noise Diffusion (CEND) layer to construct a diffusible multi-source, category-structured embedding space $\mathbf{E}$.
The synthetic images are then generated and stored in memory, with the generator network optimization details given in Section~\ref{sec:Optimization}. 
During the student network update, synthetic data from memory is fed into both the teacher and student networks. In addition to traditional logit knowledge alignment, the Category Noise Contrastive Learning~(CNCL) module encourages the student network to focus on robust domain-invariant category features beneficial to strong generalization.

\subsection{Initial Category Embedding and Noise Diffusion}
\label{Initial Category Embedding and Noise Diffusion}
Unlike native DFKD, which samples unstructured Gaussian noise as the generator embedding inputs, we employ a pre-trained language model (LM) to introduce structured initial category embeddings.  
Specifically, as illustrated in Figure~\ref{fig:f3}~\textbf{(a)}, we construct prompts of the form ``\textit{a photo of \{class\}}" as input to the LM, where ``\textit{\{class\}}" can be the class name or class index (further explored in Section~\ref{sec:limit}).  
For example, for the class ``cat", we input \textit{``a photo of cat"} into the LM, obtaining the corresponding initial category embedding $\mathbf{e}_{\text{cat}}^{\text{off}} \in \mathbb{R}^D$ where $D$ is the embedding dimension. 
This process is repeated for all $K$ categories, forming the initial category embedding space $\mathbf{E}^{\text{off}} = \{\mathbf{e}_k^{\text{off}}\}_{k=1}^K$, where $\mathbf{E}^{\text{off}} \in \mathbb{R}^{K \times D}$ and ``off" indicate that this operation takes place offline and is only performed once throughout the entire DFKD process, without introducing any additional training overhead.

\begin{figure}[!t]
    \centering
    \includegraphics[width=0.9\linewidth]{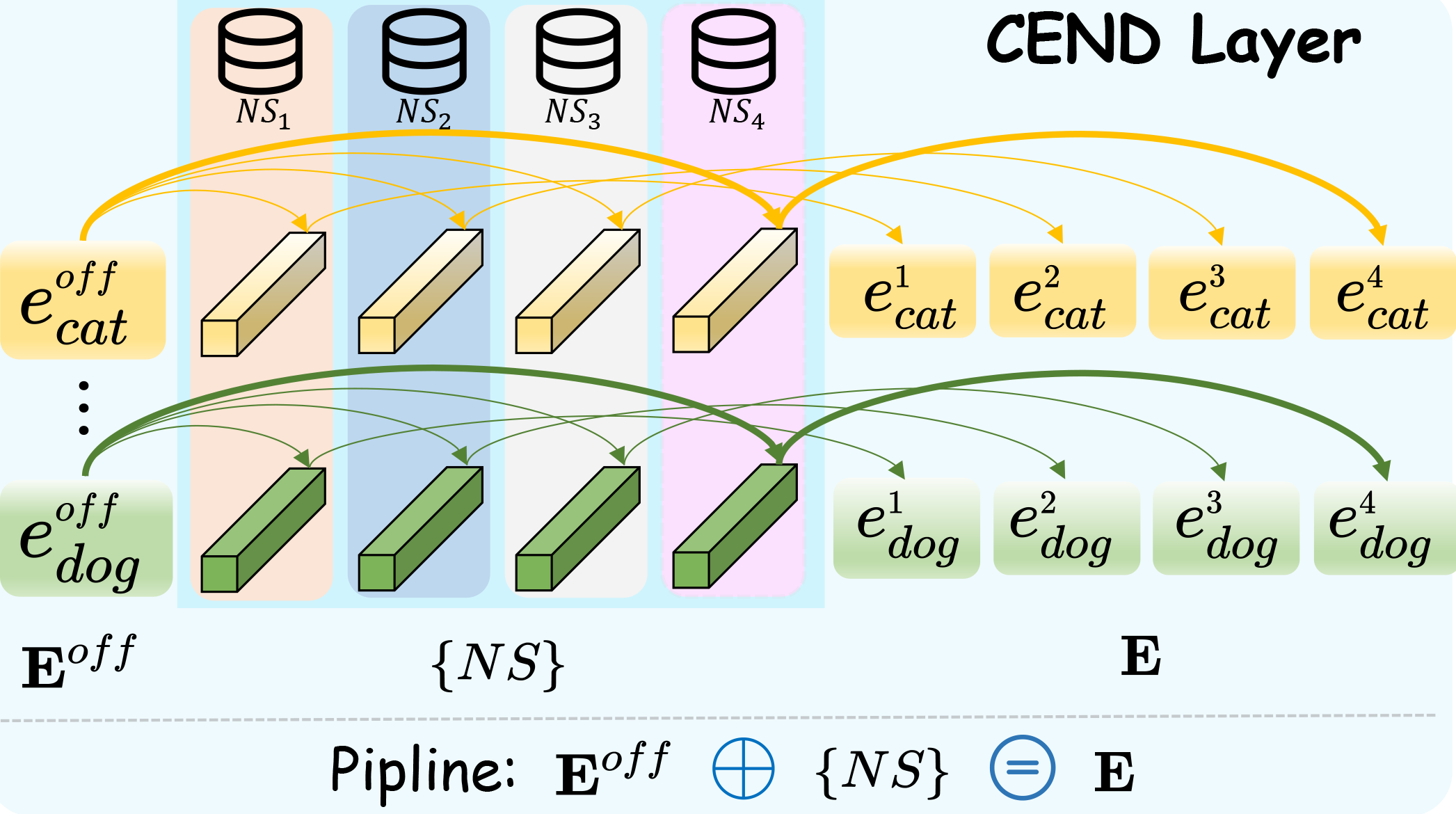} 
    \caption{\textbf{Category Embedding Noise Diffusion (CEND).}
    We leverage a pre-trained language model to initialize category embeddings ($\mathbf{E}^{\text{off}}$) with structure, in contrast to the unstructured Gaussian noise used in native DFKD.  However, $\mathbf{E}^{\text{off}}$ suffers from sparsity and lacks diversity.  
 Our proposed CEND layer introduces $N$ noise sources ($\{NS\}$), each following a distinct pre-defined distribution (illustrated here with N=4),  sampling from each $NS_n (n=1,\cdots4)$  and element-adding it to $\mathbf{E}^{\text{off}}$ enables dynamic diffusion of the initial embeddings, resulting in a richer embedding space $\mathbf{E}$.
    }
    \label{fig:f4}
    \vspace{-0.5cm}
\end{figure}

Although $\mathbf{E}^{\text{off}}$ exhibits the initial category structure compared to Gaussian noise, it remains sparse and lacks the embedding space richness.  To address this, we introduce the \textbf{Category Embedding Noise Diffusion (CEND)} layer, detailed in Figure~\ref{fig:f3}~\textbf{(b)} or Figure~\ref{fig:f4}.  
CEND takes $\mathbf{E}^{\text{off}}$ as input and samples from $N$ noise sources (illustrated with $N=4$). 
Each noise source, denoted as $\textbf{NS}_n$ for $n \in \{1, 2, \dots, N\}$, follows a distinct pre-defined noise distribution, generating the corresponding noise vectors $\mathbf{q}_n \in \mathbb{R}^D$, which guides $\mathbf{E}^{\text{off}}$ to diffuse into diverse latent embedding spaces.  
For the ``cat" category, we sample noise vectors $\mathbf{q}_n$ from each $\textbf{NS}_n$ and perform element-wise calculation to diffuse the initial category embedding $\mathbf{e}_{\text{cat}}^{\text{off}}$:
\begin{equation}
    \begin{aligned}
             \mathbf{e}_{\text{cat}}^n = \mathbf{e}_{\text{cat}}^{\text{off}} \oplus (\mathbf{M}_n \odot \mathbf{q}_n), \quad n \in \{1, 2, \dots, N\}
    \end{aligned}
\end{equation}
where $\mathbf{M}_n$ is the magnitude of noise perturbation from source $\textbf{NS}_n$, \(\oplus\) denotes the element-wise addition and  $\odot$ represents element-wise multiplication. 
The same process is applied for all $K$ categories, producing diffused embedding space $\mathbf{E}$, as shown in Figure~\ref{fig:f4}, which serves to enrich the category-structured embedding space.

\begin{table*}[!t]
\centering
\setlength{\abovecaptionskip}{-0.03cm}
    \caption{
    {Smaller resolution experiments}.
    Comparison of our \textbf{CAE-DFKD} with other data-free  methods under different configurations. The best results, signified in bold,
    and the symbols $[^{*}]$ and $[^{\lambda}]$ are used here to indicate that the results are from \cite{yu2023data}  and NAYER\cite{nayer}, respectively.
}
\resizebox{0.9\textwidth}{!}{
    \begin{tabular}{lccccccccccc}

    \toprule[1.5px] 
    {} & & \multicolumn{5}{c}{\textbf{CIFAR-100}} & \multicolumn{5}{c}{\textbf{CIFAR-10}}\\
    \cmidrule[0.5pt](rl){3-7}
    \cmidrule[0.5pt](rl){8-12}
    \multirow{1}{*}{\textbf{Methods}} & {\makecell[c]{\textbf{Data} \\ \textbf{Available}}} & \textbf{ResNet-34}  & \textbf{VGG-11} & \textbf{WRN-40-2}  & \textbf{WRN-40-2} & \textbf{WRN-40-2} & \textbf{ResNet-34}  & \textbf{VGG-11} & \textbf{WRN-40-2}  & \textbf{WRN-40-2} & \textbf{WRN-40-2}\\
    \multirow{2}{*}{} &  & \textbf{ResNet-18}  & \textbf{ResNet-18} & \textbf{WRN-16-1} & \textbf{WRN-40-1} & \textbf{WRN-16-2} & \textbf{ResNet-18}  & \textbf{ResNet-18} & \textbf{WRN-16-1}  & \textbf{WRN-40-1} & \textbf{WRN-16-2}\\
    \cmidrule{1-12}
    Teacher &  \CheckmarkBold & 78.05  & 71.32 & 75.83 & 75.83 & 75.83 &  95.70  & 92.25  &  94.87  &  94.87 &  94.87  \\
    Student &  \CheckmarkBold & 77.10   & 77.10 &  65.31 & 72.19 &  73.56 & 95.20   & 95.20  &  91.12  & 93.94 &  93.95   \\
    \cmidrule{1-12}
    DAFL$^{*}$\cite{chen2019data}$_{ICCV'19}$ & \XSolidBrush  & 74.47 & 57.29 & 22.50 & 34.66 & 40.00 & 92.22 & 81.10 & 65.71  & 81.33 & 81.55  \\
    ZSKT$^{\lambda}$\cite{micaelli2019zero}$_{NIPS'19}$ & \XSolidBrush & 67.74 & 54.31 & 36.60 & 53.60 & 54.59 & 93.32 & 89.46 & 83.74  & 86.07 & 89.66  \\
    DeepInv$^{\lambda}$\cite{yin2020dreaming}$_{CVPR'20}$ & \XSolidBrush & 61.32 & 54.13 & 53.77 & 68.58 & 61.34 & 93.26 & 90.36 & 83.04 & 86.85 & 89.72  \\
    DFQ$^{\lambda}$\cite{choi2020data}$_{CVPR'20}$ & \XSolidBrush & 77.01 & 66.21 & 51.27 & 54.43 & 64.79 & 94.61 & 90.84 & 86.14  & 91.69 & 92.01  \\
    CMI$^{\lambda}$\cite{fang2021contrastive}$_{IJCAI'21}$ & \XSolidBrush & {77.04} & {70.56} & {57.91} & {68.88} & {68.75} & {94.84} & {91.13} & {90.11}  & {92.78} & 92.52  \\
    FM$^{\lambda}$\cite{fang2022up}$_{AAAI'22}$ & \XSolidBrush & 74.34 & 67.44 & 54.02 & 63.91 & 65.12 & {94.05} & 90.53 & 89.29  & 92.51 & 92.45  \\
    SpaceShipNet$^{*}$\cite{yu2023data}$_{CVPR'23}$ & \XSolidBrush & \thirdcolor{77.41} & \thirdcolor{71.41} & {58.06} & {68.78} & {69.95} &  \firstcolor{\textbf{95.39}} & \thirdcolor{92.27} & {90.38}  & {93.56} & {93.25} \\
    SSD-KD\cite{liu2024small}$_{CVPR'24}$ & \XSolidBrush & 75.16 & 68.77 & 55.61 & 64.57 & 65.28 & 94.26 & 90.67 & 89.96  & {93.23} & {93.11} \\
    KDCI + CMI\cite{wang2024confounded}$_{CVPR'24}$ & \XSolidBrush & 75.07 & 69.07 & 57.19 & 67.47 & 67.68 & 94.43 & {91.28} & 89.52  & 92.84 & 92.73 \\
    CCL-D\cite{li2024category}$_{TMM'24}$ & \XSolidBrush & 77.12 & 71.02 & \thirdcolor{61.04} & \thirdcolor{70.21} & \thirdcolor{70.59} & \secondcolor{95.36} & 91.85 & \thirdcolor{91.76}  & \secondcolor{94.27} & \thirdcolor{94.05} \\
    NAYER \cite{nayer}$_{CVPR'24}$ & \XSolidBrush & \secondcolor{77.54} & \secondcolor{71.75} & \secondcolor{62.23} & \secondcolor{71.80} & \secondcolor{71.72} & \thirdcolor{95.21} & \secondcolor{92.37} & \secondcolor{91.94}  & \thirdcolor{94.15} & \secondcolor{94.07} \\
    \cmidrule{1-12}
    \rowcolor{gray!13} \textbf{CAE-DFKD~(Ours)} & \XSolidBrush & \firstcolor{\textbf{77.83}}  & \firstcolor{\textbf{72.36}}  & \firstcolor{\textbf{62.73}}  & \firstcolor{\textbf{72.23}}  &  \firstcolor{\textbf{72.09}} & {{95.09}}  & \firstcolor{\textbf{92.74}} & \firstcolor{\textbf{92.21}}  & \firstcolor{\textbf{94.89}}  &  \firstcolor{\textbf{94.36}}  \\
    \bottomrule[1.5px]
    \end{tabular}
}

\vspace{-0.4cm}
\label{tab:tab1}
\end{table*}

\subsection{Contrastive Pairs Construct: Image-Level $\rightarrow$ Embedding-Level}
\label{sec:cncl}
Traditional contrastive learning constructs positive-negative pairs at the image level,  however, as analyzed in Section~\ref{Introduction} (both quantitatively and qualitatively), this native design provides limited benefit to the student network in DFKD.  
We propose an embedding-level method that exploits the synthesizing ability of the generator network to indirectly construct positive-negative contrastive pairs, thereby avoiding direct comparisons among often lower-quality synthetic images.

We introduce the \textbf{Category Noise Contrastive Learning (CNCL)} module, which uses the initial category embeddings $\mathbf{E}^{\text{off}}$ and the diffused noise embeddings as generator input to produce anchor synthetic images and corresponding positive contrastive pairs.  For example, as shown in Figure~\ref{fig:f3}~\textbf{(b)}, the initial embeddings $\mathbf{e}_{\text{cat}}^{\text{off}}$ and $\mathbf{e}_{\text{dog}}^{\text{off}}$ generate anchor images $S_{\text{cat}} = G(\mathbf{e}_{\text{cat}}^{\text{off}})$ and $S_{\text{dog}} = G(\mathbf{e}_{\text{dog}}^{\text{off}})$, respectively.  The diffused embeddings $\mathbf{e}_{\text{cat}}^n$ generate $S_{\text{cat}}^n = G(\mathbf{e}_{\text{cat}}^n)$, which serve as positive pairs for $S_{\text{cat}}$. As shown in Figure~\ref{fig:f3}~\textbf{(c)}, anchor images of other categories (e.g., $S_{\text{dog}}$) and its corresponding noise-perturbed versions ($S_{\text{dog}}^n = G(\mathbf{e}_{\text{dog}}^n)$) constitute negative pairs for $S_{\text{cat}}$. 
This leads to the CNCL loss:
\begin{equation}
\mathcal{L}_{cncl} = -\frac{1}{K} \sum_{k=1}^{K} \sum_{S_{k}^+ \in \mathcal{P}_{k}} \log \frac{\exp(\text{sim}(S_k, S_k^+) / \tau)}{\sum_{S' \in \mathcal{P}_{k} \cup \mathcal{N}_{k}} \exp(\text{sim}(S_k, S') / \tau)}
\end{equation}
where $K$ is the number of categories, $\tau$ is the temperature parameter, $S_k$ is the anchor image for category $k$, $\mathcal{P}_{k} = \{S_{k}^n | n=1, \dots, N\}$ is the set of positive samples generated from the diffused embeddings of category $k$, $\mathcal{N}_{k} = \{S_{k'}^n | k' \neq k, n=1, \dots, N\}$ is the set of negative samples from other categories, and $\text{sim}(\cdot, \cdot)$ denotes the cosine similarity function.

\subsection{Optimization}
\label{sec:Optimization}

\subsubsection{\texttt{\textbf{Generator Optimization}}}
The generator network is trained using the following loss function (Figure~\ref{fig:f3}~\textbf{(b)}):
\begin{equation}
    \mathcal{L}_G = \mathcal{L}_{CE} + \lambda_{bn}\mathcal{L}_{BN} + \lambda_{adv}\mathcal{L}_{adv}
    \label{loss:g}
\end{equation}
where $\mathcal{L}_{CE}$ is the cross-entropy loss between the teacher network's predictions on synthetic images and its corresponding ground-truth labels, $\mathcal{L}_{BN}$ is a batch normalization loss for numerical stability (commonly used in DFKD), and $\mathcal{L}_{adv}$ (Eq.~\ref{eq:adv}) is the adversarial loss. The hyperparameters $\lambda_{bn}$ and $\lambda_{adv}$ balance these losses.

\subsubsection{\texttt{\textbf{Student Optimization}}}
The student network is trained using a combination of the classical knowledge distillation objective and our proposed contrastive learning method~(Figure~\ref{fig:f3}~\textbf{(c)}).  The loss function is:
\begin{equation} 
    \mathcal{L}_{S} = \mathcal{L}_{KL} + \alpha \mathcal{L}_{cncl}
\label{loss:stu}
\end{equation}
where $\mathcal{L}_{KL}$ is the KL divergence between the teacher and student logits (for knowledge distillation), $\mathcal{L}_{cncl}$ is the embedding-level category noise contrastive learning loss (introduced in Section~\ref{sec:cncl}), and $\alpha$ controls the contribution of the contrastive loss.

\section{Experiments}
\subsection{Experimental Setup}

\noindent
\textbf{Datasets:}
We comprehensively evaluate the performance of the proposed CAE-DFKD method across image recognition datasets with various input resolutions. 
For smaller resolutions of $32 \times 32$, we evaluate CAE-DFKD on CIFAR-10 and CIFAR-100, which contain 10 and 100 categories, respectively. For medium resolutions of $64 \times 64$, we employ Tiny-ImageNet, which comprises 200 categories. For larger resolutions of $224 \times 224$, we conduct evaluations on ImageNet-1K.

\noindent
\textbf{Configuration Details:}
Evaluation procedure is performed across various teacher-student combinations, including ResNet~\cite{he2016deep}, WideResNet~(WRN), and VGG~\cite{simonyan2014very} models, without access to the real dataset. 
The generator network is optimized using the objective function defined in Eq.~\ref{loss:g}, with the Adam optimizer and an initial learning rate of 0.001. For the student network, we employ the objective function defined in Eq.~\ref{loss:stu} and optimize it using the SGD optimizer with an initial learning rate of 0.1 and training epochs of $300$. 
Additionally, cosine annealing scheduling is applied to adjust the learning rate throughout the training process. All experiments are performed on the  NVIDIA RTX 3090 GPUs.


\begin{table}[!t]
\caption{ 
Medium resolution experiments. Comparison of ResNet-34 as teacher and ResNet-18 as student against SOTA methods. 
Symbol $[^{\lambda}]$ indicates that the result is from~\cite{nayer}.
}
\renewcommand{\arraystretch}{0.85} 
\label{tab:tab2}
\centering

\resizebox{0.8\linewidth}{!}{
\begin{tabular}{lcc}
\toprule[1.5px] 
\textbf{Methods} & \makecell[c]{\textbf{Data}  \textbf{Available}} & \makecell[c]{\textbf{Tiny-ImageNet} \\ \textbf{Top-1 Acc. (\%)}} \\
\cmidrule{1-3}
Teacher  & \CheckmarkBold & 66.44 \\
Student  & \CheckmarkBold   & 64.87 \\
\cmidrule{1-3}
CMI$^{\lambda}$\cite{fang2021contrastive}$_{IJCAI'21}$ & \XSolidBrush & 64.01 \\
PREKD$^{\lambda}$\cite{binici2022robust}$_{AAAI'22}$ & \XSolidBrush & 49.94 \\
MBDFKD$^{\lambda}$\cite{binici2022preventing}$_{ICCV'22}$ & \XSolidBrush & 47.96 \\
MAD$^{\lambda}$\cite{do2022momentum}$_{NIPS'22}$ & \XSolidBrush & 62.32 \\
KAKR$_{MB}$$^{\lambda}$\cite{patel2023learning}$_{CVPR'23}$ & \XSolidBrush & 47.96 \\
KAKR$_{GR}$$^{\lambda}$\cite{patel2023learning}$_{CVPR'23}$ & \XSolidBrush & 49.88 \\
SpaceShipNet\cite{yu2023data}$_{CVPR'23}$ & \XSolidBrush & \thirdcolor{64.04} \\
KDCI + DFND\cite{wang2024confounded}$_{CVPR'24}$ &  \XSolidBrush & 49.54 \\
NAYER\cite{nayer}$_{CVPR'24}$ & \XSolidBrush & \secondcolor{64.17} \\
\cmidrule{1-3}
\rowcolor{gray!13}  \textbf{CAE-DFKD~(Ours)} & \XSolidBrush & \firstcolor{\textbf{64.72}} \\
\bottomrule[1.5px] 
\end{tabular}
}
\end{table}

\begin{table}[!t]
\caption{ 
Larger resolution experiments. ResNet-50$\rightarrow$ResNet-50. 
Symbol $[^{\lambda}]$ indicates that the result is from~\cite{nayer}.
}
\label{tab:tab3}
\renewcommand{\arraystretch}{0.8} 
\centering
\resizebox{0.73\linewidth}{!}{
\begin{tabular}{lcc}
\toprule[1.5px] 
\textbf{Methods} & \makecell[c]{\textbf{Data} \textbf{Available}} & \makecell[c]{\textbf{ImageNet-1K} \\ \textbf{Top-1 Acc. (\%)}} \\
\cmidrule{1-3}
Teacher  & \CheckmarkBold & 75.45 \\
Student  & \CheckmarkBold & 75.45 \\
\cmidrule{1-3}
FM$^{\lambda}$\cite{fang2022up}$_{AAAI'22}$ & \XSolidBrush & 57.37 \\
DeepInv$^{\lambda}$\cite{yin2020dreaming}$_{CVPR'20}$ & \XSolidBrush & \thirdcolor{68.00} \\
NAYER\cite{nayer}$_{CVPR'24}$ & \XSolidBrush & \secondcolor{68.92} \\
\cmidrule{1-3}
\rowcolor{gray!13}  \textbf{CAE-DFKD~(Ours)} & \XSolidBrush & \firstcolor{\textbf{69.33}} \\
\bottomrule[1.5px] 
\end{tabular}
}

\end{table}

\begin{table*}[!t]
\small
\centering
\setlength{\abovecaptionskip}{-0.03cm}

\begin{minipage}{0.25\textwidth}
    \caption{On the \textbf{NYUv2} dataset, transferability of learned knowledge with our CAE-DFKD is evaluated.
    Student: ResNet-34.
    }
    \label{tab:nyu}
\end{minipage}%
\begin{minipage}{0.7\textwidth}
\centering
\resizebox{\linewidth}{!}{
    \begin{tabular}{lcccccccccc}
    \toprule[1.5px]
    {} &  & \multicolumn{2}{c}{\textbf{Semantic Segmentation}} & \multicolumn{2}{c}{\textbf{Depth Estimation}} & \multicolumn{5}{c}{\textbf{Surface Normal Prediction}} \\
    \cmidrule[0.5pt](rl){3-4}
    \cmidrule[0.5pt](rl){5-6}
    \cmidrule[0.5pt](rl){7-11}
    \multirow{2}{*}{\textbf{Method}} & \multirow{2}{*}{\textbf{Data Available}} & \multirow{2}{*}{{mIoU $\uparrow$}}  & \multirow{2}{*}{{pAcc~(\%) $\uparrow$}}  & \multirow{2}{*}{{AErr $\downarrow$}}   & \multirow{2}{*}{{RErr $\downarrow$}}  & \multicolumn{2}{c}{{Angle Distance}} & \multicolumn{3}{c}{{Within $t^{\circ}$}} \\
    \cmidrule[0.5pt](rl){7-8}
    \cmidrule[0.5pt](rl){9-11}
     & &  &   &  &  & {Mean $\downarrow$} & {MED $\downarrow$} & {11.25 $\uparrow$} & {22.5 $\uparrow$} & {30 $\uparrow$} \\
    \midrule
    Teacher & \CheckmarkBold & 37.61  & 64.03  &  0.5137  &  0.2194  & 27.82  &  20.92  & 28.99  &   52.78 & 64.13  \\
    \midrule
    Student & \CheckmarkBold & \firstcolor{34.09}  & \firstcolor{61.28}  & \secondcolor{0.5349}   & \firstcolor{0.2219} & \firstcolor{29.68}  & \firstcolor{24.16}   & \secondcolor{26.77} & \firstcolor{50.14}   & \firstcolor{59.94} \\
    NAYER\cite{nayer} & \XSolidBrush &  \thirdcolor{31.82}  & \thirdcolor{59.12}  & \thirdcolor{0.5678}   & \thirdcolor{0.2535}  & \thirdcolor{32.18}  & \thirdcolor{27.35}   & \thirdcolor{26.14} & \thirdcolor{46.28}   & \thirdcolor{57.68} \\
     \rowcolor{gray!20} \textbf{CAE-DFKD~(Ours)} & \XSolidBrush & \textbf{\secondcolor{33.77}}    & \textbf{\secondcolor{60.86}}   & \textbf{\firstcolor{0.5211}}    & \textbf{\secondcolor{0.2374}}  & \textbf{\secondcolor{30.06}}   & \textbf{\secondcolor{24.99}}    & \textbf{\firstcolor{27.36}} & \textbf{\secondcolor{48.33}}    & \textbf{\secondcolor{58.16}}  \\
    \bottomrule[1.5px]
    \end{tabular}
}
\end{minipage}
\vspace{-0.2cm}
\end{table*}

\begin{table*}[!t]
\centering
\setlength{\abovecaptionskip}{-0.03cm}

\begin{minipage}{0.25\textwidth}
    \caption{Validate the effectiveness of CAE-DFKD in transfer learning on the \textbf{ADE-20K} and \textbf{COCO-2017} datasets.
    Student: ResNet-34.
    }
    \label{tab:coco}
\end{minipage}%
\begin{minipage}{0.7\textwidth}
\centering
\resizebox{\linewidth}{!}{
\begin{tabular}{lccccccccc}
\toprule[1.5px]
\multirow{2}{*}{\textbf{Methods}} & \multirow{2}{*}{\textbf{Data Available}} & \multicolumn{2}{c}{\textbf{ADE-20K}} & \multicolumn{6}{c}{\textbf{COCO-2017}} \\
\cmidrule[0.5pt](rl){3-4}
\cmidrule[0.5pt](rl){5-10}
                         & & pAcc~(\%)$\uparrow$ & mIoU$\uparrow$ & mAP$^{_{bbox}}\uparrow$ & mAP$^{_{bbox}}_{50}\uparrow$ & mAP$^{_{bbox}}_{75}\uparrow$ & mAP$^{_{bbox}}_s\uparrow$ & mAP$^{_{bbox}}_m\uparrow$ & mAP$^{_{bbox}}_l\uparrow$ \\
\midrule
Teacher & \CheckmarkBold & 82.49 & 39.34 & 34.08 & 52.91 & 35.42 & 19.85 & 37.51 & 46.34 \\
\midrule
Student & \CheckmarkBold & \firstcolor{78.02} & \firstcolor{36.91} & \firstcolor{31.00} & \firstcolor{48.90} & \firstcolor{32.70} & \secondcolor{15.30} & \secondcolor{33.50} & \firstcolor{43.10} \\
CMI~\cite{fang2021contrastive} & \XSolidBrush & 74.11 & 33.72 & \thirdcolor{29.88} & 45.33 & \thirdcolor{31.04} & 13.24 & 29.15 & 38.47 \\
SpaceShipNet~\cite{yu2023data} & \XSolidBrush & \thirdcolor{75.87} & \thirdcolor{34.09} & 29.47 & \thirdcolor{46.90} & 30.60 & \thirdcolor{13.52} & \thirdcolor{31.29} & \thirdcolor{41.06} \\

\rowcolor{gray!20} \textbf{CAE-DFKD~(Ours)} & \XSolidBrush & \textbf{\secondcolor{77.14}} & \textbf{\secondcolor{35.66}} & \textbf{\secondcolor{30.43}} & \textbf{\secondcolor{48.26}} & \textbf{\secondcolor{32.18}} & \textbf{\firstcolor{16.79}} & \textbf{\firstcolor{34.43}} & \textbf{\secondcolor{42.35}} \\
\bottomrule[1.5px]
\end{tabular}
}
\end{minipage}
\vspace{-0.4cm}
\end{table*}

\subsection{Main Results}
\label{main result}

\subsubsection{\texttt{\textbf{Comparison with State-of-the-art Methods}}}
Tables \ref{tab:tab1}, \ref{tab:tab2}, and \ref{tab:tab3} present the  experimental results\footnote{Top three methods in descending order: \firstcolor{$1_{st}$}, \secondcolor{$2_{nd}$}, \thirdcolor{$3_{rd}$}} conducted at progressively increasing resolutions: smaller~($32 \times 32$), medium~($64 \times 64$), and larger~($224 \times 224$), respectively. 
``Teacher" and "Student" denote the results obtained using the data-accessible setting.
In Table~\ref{tab:tab1}, focusing on the smaller resolution corresponding to CIFAR-10 and CIFAR-100, CAE-DFKD consistently outperforms other methods, except for the ResNet-34 $\rightarrow$ ResNet-18 combination on CIFAR-10.  
At medium resolution for Tiny-ImageNet, most current DFKD methods\cite{nayer,yu2023data,binici2022robust} select the ResNet-34 $\rightarrow$ ResNet-18 setting for testing, as shown in Table~\ref{tab:tab2}, where our CAE-DFKD surpasses SOTA NAYER~\cite{nayer}~($64.17\% \rightarrow 64.72\%$). 
In Table~\ref{tab:tab3} at the larger resolution for ImageNet-1K, due to efficiency constraints, existing DFKD methods rarely report results.
We follow the classic ResNet-50 $\rightarrow$ ResNet-50 setting\cite{nayer,yin2020dreaming,fang2022up}, using PyTorch-provided pre-trained weights for the teacher ResNet-50.





\subsubsection{\texttt{\textbf{Generalization Performance Comparison}}}
\label{sec:transfer}
We conduct extensive experiments to validate the transferability of learned representations  by student networks.
Specifically, we perform the evaluation under both data-accessible and data-free settings across a range of downstream tasks, including semantic segmentation, depth estimation, surface normal prediction, and object detection. 
The evaluation metrics for each task are detailed as follows:
\begin{itemize}
    \item {NYUv2}:  A benchmark for scene understanding, NYUv2 employs distinct metrics for each subtask.  Semantic segmentation evaluation via Mean Intersection over Union (mIoU) and Pixel Accuracy (pAcc). 
    Depth estimation evaluation using absolute and relative error. For surface normal estimation, we consider the mean absolute error (Mean), median absolute error (Median), and percentages of pixels with angular errors within 11.25, 22.5, and 30 degrees. 
    \item {ADE-20K}:   Serving as a benchmark~\cite{zhou2017scene} to evaluate scene parsing capabilities, relies on semantic segmentation and employs pAcc and mIoU, consistent with NYUv2 evaluation. Training and evaluation are conducted using the open-source framework. 
    \item {COCO-2017:} A large-scale image recognition dataset, COCO-2017~\cite{lin2014microsoft} includes  various subtasks.  
    We focus on evaluating transfer capabilities within the object detection subtask. Metrics include the mean Average Precision (mAP) at different IoU thresholds (50, 75) and  detection behavior across object sizes, categorized as small (s), medium (m), and large (l).
\end{itemize}
In addition,  CAE-DFKD and the other compared methods  perform complete data-free knowledge distillation on CIFAR-100, and the obtained Student weights are then \textbf{fine-tuned} for downstream tasks.

Tables~\ref{tab:nyu}
and~\ref{tab:coco} present the evaluation results for downstream tasks, comparing the proposed CAE-DFKD with state-of-the-art (SOTA) DFKD methods like CMI~\cite{fang2021contrastive}, SpaceShipNet~\cite{yu2023data}, and NAYER~\cite{nayer},  ``Student" denotes data-accessible settings.  
Although SOTA DFKD methods might achieve comparable or marginally superior accuracy against the data-accessible setting on the real image recognition dataset, this advantage does not consistently extend to broader tasks, falling short of the data-accessible ``Student". 
Notably, our method, in addition to maintaining comparable performance on the real image recognition dataset, prioritizes learning generalizable and transferable category representations.  
CAE-DFKD not only exceeds existing SOTA DFKD methods, but also outperforms data-accessible ``Student" settings in certain scenarios, including absolute error~(AErr) in depth estimation, the 11.25$^{\circ}$ setting in surface normal prediction, and COCO-2017 with small and medium object~(mAP$^{_{bbox}}_s$, mAP$^{_{bbox}}_m$). 



\begin{figure*}[!t]
\setlength{\abovecaptionskip}{-0.1cm}
\begin{center}
    \begin{minipage}{0.75\textwidth} 
        \includegraphics[width=\linewidth]{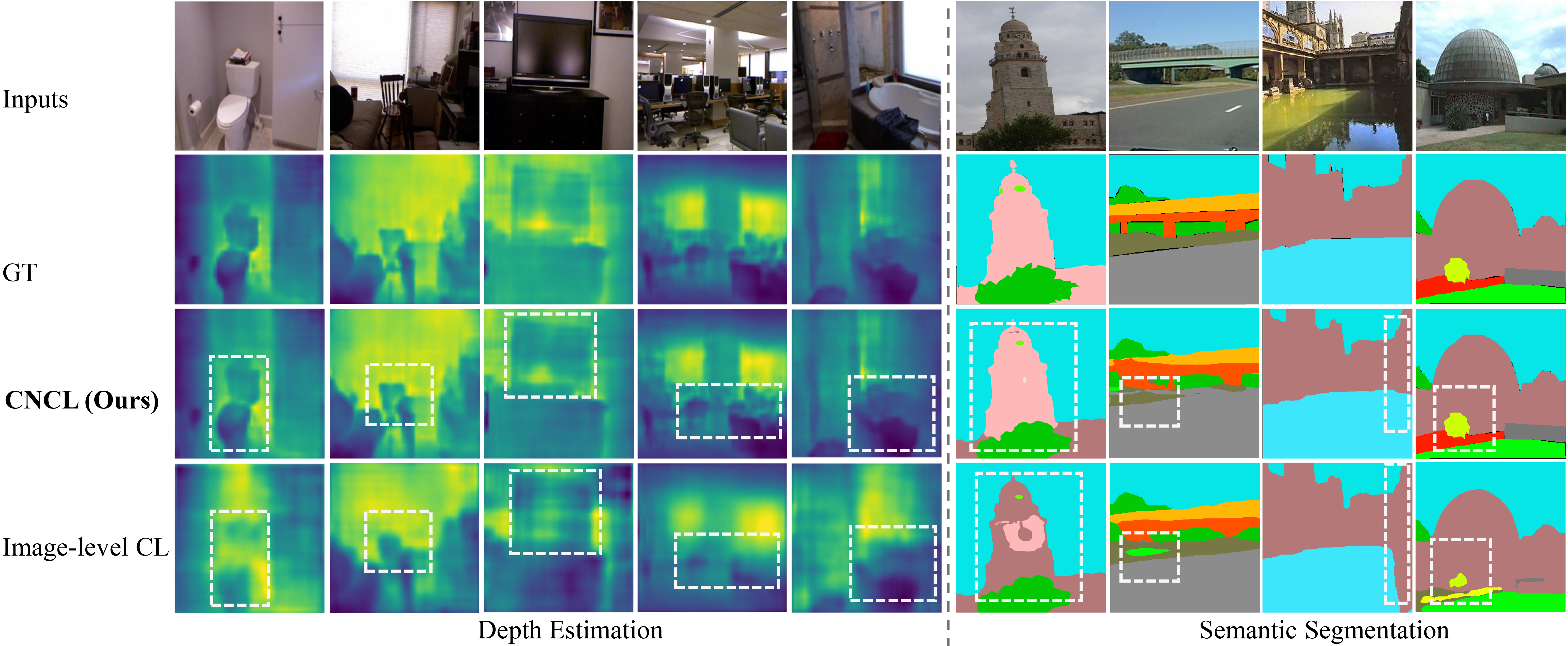}
    \end{minipage}
    \begin{minipage}{0.19\textwidth} 
        \vspace{0pt}  
        \caption{\textbf{Downstream Task Performance Visualization Comparison}.
        The contrastive learning designed for image-level optimization proves to be less effective in enhancing the generalization ability of the student network in DFKD (Section~\ref{Introduction}). In contrast, our embedding-level contrastive learning encourages the student network to learn domain-invariant class features, which benefits both depth estimation and semantic segmentation.
        }
        \label{fig:f5}
    \end{minipage}
\end{center}
\vspace{-0.6cm}
\end{figure*}

\section{ABLATION EXPERIMENTS}
\label{Ablation Experiments}


\subsubsection{\texttt{\textbf{Ablation Study of CEND and CNCL}}}
Table~\ref{tab:tab7} uses CMI\cite{fang2021contrastive} as the base method to validate the effectiveness of our proposed CEND layer and CNCL module under the data-free setting.  
It is observed that the category-structured embeddings introduced by CEND, compared to the un-structured Gaussian noise embeddings commonly used in DFKD, introduce embedding-level richness and alleviate the convergence burden of the generator, which helps produce high-quality synthetic images. 
Additionally, CNCL constructs positive-negative contrastive pairs at the embedding level to mitigate the issue of significant difference in synthetic image quality, and encourages the student network to focus on category representations with strong domain generalization, thereby benefiting downstream tasks, as shown in Figure~\ref{fig:f5}.




\begin{table}[!t]
\centering
\renewcommand{\arraystretch}{0.8}
\caption{Ablation study on \textbf{CAE-DFKD} components. Base method: CMI\cite{fang2021contrastive}. Dataset: ADE-20K, follow setting in Section~\ref{sec:transfer}. }
\resizebox{0.85\linewidth}{!}{
\begin{NiceTabular}{l|ccc|ll} 
        \toprule[1.5px] 
        {\textbf{Methods}} & $\mathcal{L}_{KD}$ & {\textbf{CEND}} & {\textbf{CNCL}} & {\textbf{pAcc.} (\%)} & {\textbf{mIoU}} \\
        \cmidrule{1-6}
        \multicolumn{6}{c}{\emph{ResNet-34 $\rightarrow$ ResNet-18}} \\
        Base & \CheckmarkBold &  &  & 74.11$^{\mycolor[0.9]{\mathbf{0.0}}}$ & 33.72$^{\mycolor[0.9]{\mathbf{0.0}}}$ \\
        \rowcolor{gray!13} \multirow{2}{*}{\textbf{+ CAE-DFKD}}   
            & \CheckmarkBold & \CheckmarkBold  &  & 75.38$^{\mycolor[0.9]{\mathbf{+1.27}}}$ &  34.48$^{\mycolor[0.9]{\mathbf{+0.76}}}$ \\ 
         \rowcolor{gray!13} 
            & \CheckmarkBold & \CheckmarkBold & \CheckmarkBold & \textbf{77.14}$^{\mycolor[0.9]{\mathbf{+3.03}}}$ & \textbf{35.66}$^{\mycolor[0.9]{\mathbf{+1.94}}}$  \\
         \hline
         \midrule
         \multicolumn{6}{c}{\emph{WRN-40-2 $\rightarrow$ WRN-40-1}} \\
        Base & \CheckmarkBold  &  &  & 67.52$^{\mycolor[0.9]{\mathbf{0.0}}}$ & 29.47$^{\mycolor[0.9]{\mathbf{0.0}}}$ \\
       \rowcolor{gray!13} \multirow{2}{*}{\textbf{+ CAE-DFKD}}   
            & \CheckmarkBold  & \CheckmarkBold  &   & 70.31$^{\mycolor[0.9]{\mathbf{+2.81}}}$ & 30.97$^{\mycolor[0.9]{\mathbf{+1.50}}}$  \\ 
         \rowcolor{gray!13} 
            & \CheckmarkBold  & \CheckmarkBold & \CheckmarkBold &  \textbf{72.49}$^{\mycolor[0.9]{\mathbf{+4.97}}}$ & \textbf{33.08}$^{\mycolor[0.9]{\mathbf{+3.61}}}$  \\ 
        \bottomrule[1.5px] 
        \end{NiceTabular}
}
\label{tab:tab7}
\end{table}

\begin{table}[!t]
\setlength{\abovecaptionskip}{-0.03cm}
\caption
{
Ablation study on the  noise disturbances \(N\) for category embeddings. 
NYAER\cite{nayer} as the base method, follow setting in Section~\ref{sec:transfer}.
Dataset: NYUv2 Semantic Segmentation~(mIoU).
}

\centering

\resizebox{0.95\linewidth}{!}{
\setlength{\tabcolsep}{1.3mm}{
\begin{NiceTabular}{l|c|c|c|c|c}
        \toprule[1.5px] 
        \multirow{2}{*}{Methods}  & \multicolumn{5}{c}{\textbf{Effect of the NS Number}~($N$)}   \\
        \cline{2-6}
         & 2  & 3 & \cellcolor{gray!15} \textbf{4} & 5  & 6  \\
        \midrule
        \multicolumn{6}{c}{\emph{{ResNet-34} $\rightarrow$ {ResNet-18}}} \\
        Base & \multicolumn{5}{c}{31.82$^{\mycolor[0.9]{\mathbf{0.0}}}$} \\ 
        \rowcolor{gray!13}   \textbf{+ CAE-DFKD} & 32.14$^{\mycolor[0.9]{\mathbf{+0.32}}}$ & 32.39$^{\mycolor[0.9]{\mathbf{+0.57}}}$ & \textbf{33.77}$^{\mycolor[0.9]{\mathbf{+1.95}}}$ & 33.29$^{\mycolor[0.9]{\mathbf{+1.47}}}$ & 33.01$^{\mycolor[0.9]{\mathbf{+1.19}}}$ \\ 
        \hline
        \hline
        \multicolumn{6}{c}{\emph{{WRN-40-2} $\rightarrow$ {WRN-40-1}}} \\
        Base & \multicolumn{5}{c}{26.84$^{\mycolor[0.9]{\mathbf{0.0}}}$} \\ 
        \rowcolor{gray!13}  \textbf{+ CAE-DFKD} & 27.24$^{\mycolor[0.9]{\mathbf{+0.40}}}$ & 27.28$^{\mycolor[0.9]{\mathbf{+0.44}}}$  & \textbf{29.28}$^{\mycolor[0.9]{\mathbf{+2.44}}}$  & 28.17$^{\mycolor[0.9]{\mathbf{+1.33}}}$  & 28.87$^{\mycolor[0.9]{\mathbf{+2.03}}}$   \\ 
        \bottomrule[1.5px] 
    \end{NiceTabular}
}
}
\label{tab:tab8}
\vspace{-0.3cm}
\end{table}

\subsubsection{\texttt{\textbf{Effect of Noise Sources Number~($N$) on Generalization}}}
In CEND, we introduce the noise diffusion method into the initial category embeddings to enrich the embedding space, which is influenced by the noise diffusion source number $N$. 
In addition, $N$ also affects the construction of positive-negative contrastive pairs in CNCL. In Table~\ref{tab:tab8}, we present the impact of different $N$ on the performance of the student network in downstream tasks. 
When $N$ is relatively small ($<4$), CEND struggles to introduce sufficient richness to the embedding space $\mathbf{E}$. 
In contrast, when $N$ becomes too large ($>4$), an excessive number of positive and negative pairs is generated for CNCL, which decreases training efficiency and increases the difficulty of semantic contrast for noisy images, thus placing a heavier burden on the optimization process of the student network. Despite this, our CAE-DFKD significantly outperforms the base method under different values of $N$, with the most robust performance achieved when $N = 4$.

\subsubsection{\texttt{\textbf{Efficiency Advantage of CEND}}}
Under native DFKD generator network update setting, which samples Gaussian noise and is required to project into highly structured dataset structures (``\textbf{un-structured $\rightarrow$ structured}"), faces extreme difficulties, particularly in the data-free setting where no real images are available to provide discriminative information. In contrast, our CAE-DFKD approach introduces rich structured embeddings through CEND, transforming the generator update process into ``\textbf{structured $\rightarrow$ structured}", significantly alleviating the convergence burden on the generator. Table~\ref{tab:tab9} reports the efficiency gains made by CEND.

\subsubsection{\texttt{\textbf{LMs on Generalization}}}
In Section~\ref{Initial Category Embedding and Noise Diffusion}, we employ a pre-trained LM to provide structured category information for the generator embedding inputs. 
In Table~\ref{tab:tab10}, we report the impact of different LMs on the downstream task of COCO-2017 object detection. It can be observed that, compared to the base, the specific choice of LM (doc2vec, CLIP, SBERT) shows little sensitivity, as all lead to significant performance improvements. However, CLIP yields superior performance.


\begin{table}[!t]
\small
\setlength{\abovecaptionskip}{-0.03cm}
\caption
{
Training Epoch Time Comparison in Minutes~($m$).
}

\centering

\resizebox{0.95\linewidth}{!}{
\begin{NiceTabular}{l|cc}
        \toprule[1.5px] 
        \multirow{2}{*}{\textbf{Methods}}  & \multicolumn{2}{c}{\textbf{T $\rightarrow$ S}}   \\
        \cmidrule{2-3}
         & ResNet-34 $\rightarrow$ ResNet-18  & WRN-40-2 $\rightarrow$ WRN-16-1   \\
        \midrule
          CAE-DFKD \textbf{w/o} CEND & 2.76m &  3.15m \\
          \rowcolor{gray!13} \textbf{CAE-DFKD w/ CEND} & \textbf{2.01m} &  \textbf{1.84m}  \\
          \midrule
          \rowcolor{gray!13} \textbf{SpeedUp} & \textbf{1.37}$\times$ &  \textbf{1.71}$\times$ \\
        \bottomrule[1.5px] 
    \end{NiceTabular}
}

\label{tab:tab9}
\vspace{-0.3cm}
\end{table}


\begin{table}[!t]
\small
\setlength{\abovecaptionskip}{-0.03cm}
\caption
{
Effect of LMs on Generalizability, follow setting in Section~\ref{sec:transfer}. Dataset: COCO-2017, $mAP^{_{bbox}}_{50}$.
}

\centering

\resizebox{0.85\linewidth}{!}{
\setlength{\tabcolsep}{3.8mm}{
\begin{NiceTabular}{l|ccc}
        \toprule[1.5px] 
        \multirow{2}{*}{$T \rightarrow S$}  & \multicolumn{3}{c}{\textbf{Impact of LMs}}   \\
        \cmidrule{2-4}
         & doc2vec  & \cellcolor{gray!15} \textbf{CLIP} & SBERT  \\
        \midrule
          {ResNet-34} $\rightarrow$ {ResNet-18} & 47.97 &  \cellcolor{gray!15} \textbf{48.26} &  {47.28} \\
          {WRN-40-2} $\rightarrow$ {WRN-40-1} & 44.37 &  \cellcolor{gray!15} \textbf{44.78} &  {43.12} \\
        \bottomrule[1.5px] 
    \end{NiceTabular}
}}

\label{tab:tab10}
\vspace{-0.4cm}
\end{table}

\subsubsection{\texttt{\textbf{Limitation}}}
\label{sec:limit}
Data-free settings involve the absence of the real dataset. 
In scenarios demanding heightened privacy, access to class names might be restricted, or pre-trained language models may lack corresponding category priors. 
Unlike the discussion in Section~\ref{Initial Category Embedding and Noise Diffusion}, which exploits \textit{``a photo of \{class name\}"} as the default prompt, we explore a more general prompt design: \textit{"a photo of \{class index\}"}. 
The experimental results in Table~\ref{tab:prompt} demonstrate the effectiveness of both prompt designs, with the more general design exhibiting a slight negative impact, but it still demonstrates generalization advantages over other SOTA methods.

\begin{table}[htbp]
\small
\caption
{
Exploring the impact of various prompts, follow setting in Section~\ref{sec:transfer}.
Dataset: NYUv2 Semantic Segmentation~(mIoU).
}

\centering

\resizebox{0.9\linewidth}{!}{
\begin{tabular}{lccc}
        \toprule[1.5px] 
        \multirow{2}{*}{T $\rightarrow$ S} & \multirow{2}{*}{\textbf{Prompt}} & \multicolumn{2}{c}{\textbf{Semantic Segmentation}} \\
        \cmidrule{3-4}
         & & mIoU $\uparrow$ & pAcc~(\%) $\uparrow$  \\
        \midrule
   \rowcolor{gray!13}  \cellcolor{white}        \multirow{2}{*}{ResNet-34 $\rightarrow$ ResNet-18} & \textit{"a photo of \{class name\}"} & \textbf{33.77} &  \textbf{60.86}  \\
         & \textit{"a photo of \{class index\}"}  &  33.16 & 60.41\\ 
         \midrule
          \rowcolor{gray!13}  \cellcolor{white} \multirow{2}{*}{VGG-11 $\rightarrow$ ResNet-18} & \textit{"a photo of \{class name\}"} & \textbf{31.38} &   \textbf{57.76}  \\
         & \textit{"a photo of \{class index\}"}  &  31.04 & 57.22\\ 
        \bottomrule[1.5px] 
    \end{tabular}
}
\label{tab:prompt}
\vspace{-0.4cm}
\end{table}

\section{CONCLUSION}
In this paper, we propose \textbf{CAE-DFKD}, which rethinks the existing DFKD framework from the embedding-level view.
For technical contributions: \textit{\textbf{i)}} CAE-DFKD incorporates a category-aware prior, enriching the structured embedding space and alleviating the generator convergence burden;
\textit{\textbf{ii)}} Through embedding-level contrastive learning, CAE-DFKD addresses the ineffectiveness of image-level generalization methods for DFKD. 
For application contributions: CAE-DFKD extends the data-free knowledge transfer potential beyond widely focused image recognition to more diverse downstream tasks. Extensive experiments demonstrate the superiority and flexibility of CAE-DFKD.

\bibliographystyle{IEEEtran}
\bibliography{ref}

\end{document}